\pdfoutput=1

\documentclass[11pt]{article}

\usepackage[]{emnlp2021}

\usepackage{times}
\usepackage{latexsym}

\usepackage[T1]{fontenc}

\usepackage[utf8]{inputenc}

\usepackage{microtype}

 \usepackage[pdftex]{graphicx}

\title{Word-Free Spoken Language Understanding for Mandarin-Chinese}

\author{
  Zhiyuan Guo, Yuexin Li, Guo Chen, Xingyu Chen, Akshat Gupta\\
  Language Technologies Institute\\
  Carnegie Mellon University\\
  Pittsburgh, PA 15213 \\
  {\tt \{zguo2, yuexinli, gchen2, xingyuch\}@cs.cmu.edu, akshatgu@andrew.cmu.edu}
 }

\begin{document}
\maketitle
\begin{abstract}
 Spoken dialogue systems such as Siri and Alexa provide great convenience to people's everyday life. However, current spoken language understanding (SLU) pipelines largely depend on automatic speech recognition (ASR) modules, which require a large amount of language-specific training data. In this paper, we propose a Transformer-based SLU system that works directly on phones. This acoustic-based SLU system consists of only two blocks and does not require the presence of ASR module. The first block is a universal phone recognition system, and the second block is a Transformer-based language model for phones. We verify the effectiveness of the system on an intent classification dataset in Mandarin Chinese.
\end{abstract}

\section{Introduction}

Spoken dialogue systems such as Siri and Alexa have demonstrated their effectiveness in various 
spoken language processing tasks including intent recognition and slot filtering. They provide great convenience to people's everyday life. However, current spoken language understanding (SLU) systems face the challenge of insufficient annotated natural speech data. Specifically, a typical SLU pipeline consists of an automatic speech recognition (ASR) module followed by a natural language understanding (NLU)  module. The ASR module requires a large amount of labeled data for each language and thus forms a bottleneck.

\citet{akshat2021} proposes a novel system that replaces the ASR module with Allosaurus, which is a universal phone recognition system that provides phonetic transcript for the input audio. Inspired by this acoustic-only approach, we propose an acoustic-based SLU system using the Transformer model. Our SLU system is similar to theirs in that we both train our systems on phonetic transcriptions and does not depend on any data-demanding language-specific modules such as Automatic Speech Recognition (ASR) module. However, while their work converts text-based data to synthetic speech in order to generate the phonetic transcriptions, our work operates on the natural speech data directly.
To the best of our knowledge, our work is the first attempt that uses discovered phonetic units from natural speech instead of synthesized speech. Meanwhile, we use a transformer-based architecture instead of a CNN+LSTM based one in our NLU module.
We then verify the effectiveness of our Transformer-based SLU system on a small intent classification dataset in Mandarin Chinese. We conclude that our Transformer-based Spoken Language Understanding (SLU) system has slightly worse performance on phone input than the state-of-the-art architecture.

\section{Related Work}

Nowadays, Spoken Language Understanding (SLU) systems, an emerging field in between the 
areas of speech processing and natural language processing, have been widely used to process 
spoken utterances for various downstream tasks. Current researches in high resourced languages focus on building end-to-end (E2E) SLU systems \citep{Radfar2020, Lugosch2019} to eliminate errors through the pipeline. 
Traditional SLU pipelines include an automatic speech recognition (ASR) module followed by a natural language understanding (NLU) module. The language-specific ASR module converts spoken utterance into textual transcription, which is then fed into the NLU module for downstream tasks. Training ASR often requires a large amount of language-specific data and therefore poses an obstacle for applying traditional SLU pipeline on low-resourced languages.

In our work, we replace the ASR module in the traditional SLU pipeline with Allosaurus \citep{allosaurus2020}, a universal phone recognition system. Allosaurus provides language-independent and fine-grained phonetic transcriptions for input natural speech, which enables it to generalize well to low-resourced languages and novel sounds \citep{akshat2021, gupta2020mere}. In our system, we feed the phonetic transcriptions by Allosaurus to a Transformer-based language model, which is trained to understand the phonetic transcription and perform downstream tasks like intent classification.

To the best of our knowledge, there have been limited attempts to use the phonetic units for intent classification in a supervised fashion. While the previous work \citep{akshat2021} proposed an architecture based on CNNs (convolutional neural network) and LSTMs (long-short term memory), we present the first attempt to combine Transformer-based architecture with intent recognition task using discovered 
phonetic units from natural speech. In addition, we use a BERT-based pre-training method called RoBERTa \citep{liu2019roberta}. The Transformer-based architecture \citep{Trans2017} provides a better-structured memory for handling long-term dependencies in contents, resulting in robust performance 
across diverse downstream language understanding tasks; and the BERT-based pre-training procedure \citep{devlin-etal-2019-bert} gives bidirectional contextual representations of the phonetic units. 

\section{Model Description}

\subsection{Architecture}

The main part of our model is a multi-layer Transformer Encoder \citep{Trans2017}. Given an input sequence of phones
$x = (x_1, ..., x_T)$, the encoder generates a sequential representation $z = (z_1, ..., z_T)$.
The encoder consists of multiple layers, where each layer is composed of two modules:
(a) a multi-head attention module and (b) fully-connected feed-forward network. 
The attention mechanism features 
"Scaled Dot-Product Attention", which involves a query matrix $Q$, a key matrix $K$, and a value matrix $V$:
\begin{equation}
    \mathrm{Attention}(Q, K, V) = \mathrm{softmax}\left(\frac{QK^T}{\sqrt{d^k}}\right) V
\end{equation}
where $Q \in {\rm I\!R}^{d_k}$, $K \in {\rm I\!R}^{d_k}$, and 
$V \in {\rm I\!R}^{d_k}$ are the input to the attention layer.
The multi-head attention module combines outputs (i.e., heads) 
from $h$ different attention layers, with different linear projections of $Q$, $K$,
and $V$. Each head is computed as follows:
\begin{equation}
    \mathrm{head_i} = \mathrm{Attention}\left(QW_i^Q, KW_i^K, VW_i^V\right)
\end{equation}
The heads are then concatenated and projected again to form the final values of 
the multi-head attention module. The output of the multi-head attention module is then fed into a
feed-forward network, consisting of two linear layers. Between the two linear layer is
defined a ReLU activation:
\begin{equation}
    \mathrm{FFN}(x) = \mathrm{ReLU}(xW_1 + b_1)W_2 + b_2
\end{equation}

The encoder layers are stacked together, and the final output is fed into a linear layer
followed by a classifier.

\subsection{RoBERTa}

We choose RoBERTa, which uses the same transformer encoder architecture, as our pretraining approach. Compared to BERT, RoBERTa adjusts some design choices and training strategies of BERT to achieve better downstream task performance: RoBERTa replaces static masking in BERT with dynamic masking, which allows more variability in the masking pattern; RoBERTa also removes next sentence prediction from its objectives during pre-training, as it has been found to degrade the overall performance by the authors of RoBERTa. 

\subsection{Overall System}
Overall, our system consists of Allosaurus and a NLU module. Allosaurus first converts input audio into phonetic transcript. The NLU module, which is based on the Transformer, operates on the phonetic transcript to perform downstream tasks like intent classification. Our system is first pre-trained on the phonetic transcripts using RoBERTa.

\subsection{Baseline}

To establish performance reference, we use the model of an acoustic-only SLU system in the previous work: CNN + LSTM \citep{akshat2021}. Specifically, Allosaurus is used to convert natural speech audio to International Phonetic 
Alphabet (IPA) symbols (phones); input IPA symbol sequence is then passed to an embedding vector to generate vector representation, followed by two convolutional layers of 
kernel size 3 and 5; outputs of the covolutional layers are fed into a batch normalization layer, followed by a ReLU activation, and are then concatenated and fed to a unidirectional one-layer LSTM with a hidden dimension of 128. The final hidden state of LSTM is then processed through a linear layer followed by a classifier.

\section{Evaluation Metrics}

We evaluate our model with two metrics: accuracy and F1-score,

\begin{equation}
    \mathrm{Accuracy} = \frac{n_c}{n}
\end{equation}
where $n_c$ is the number of correct classifications, $n$ is the number of total classifications,

\begin{equation}
\mathrm{F1} = \frac{TP}{TP + 0.5(FP + FN)}
\end{equation}
where TP, FP, FN are the number of true positives, false positive, and false negative, respectively. Because of the imbalance of the intent classification dataset, we report the macro-averaged F1-score (the average of F1-scores for all classes) in this paper. We also present a detailed discussion on the selected datasets in Section \ref{sec:dataset}.

\section{Dataset}

\label{sec:dataset}

In our system, there are two stages of training: pre-training and fine-tuning. For the pre-training stage, we use THCHS-30 dataset released by Tsinghua University \citep{THCHS30_2015}; after pre-training, we fine-tune the model to perform intent classification on the dataset from the first CATSLU Challenge \citep{catslu2019}.

The CATSLU dataset contains speech recordings and text transcripts, annotated with semantic intent labels, between users (humans) and service assistants (robots).  To obtain finite number of labels and phonetic forms of input, we conduct the following steps to prepare the data:

1.	For each text transcript in the dataset, if its corresponding audio recording is available, input it raw audio clip into Allosaurus for phonetic transcripts; otherwise, discard it.

2.	Map each semantic intent annotation to numeric class labels. For example, "DecreaseBrightness" = 0.

3.	Review the most common labels as well as the balance of labels within each dataset, downsampling/upsampling until reasonable balance is achieved.

\subsection{Pre-training dataset: THCHS-30}
We attempted to improve the performance on intent classification task by pre-training the model using masked language modeling on reasonable amount of speech corpus. We use THCHS-30 dataset released by Tsinghua University. 
The pre-training dataset contains natural speech audios with varying length. There are approximately 35 hours of speech data in THCHS-30.

\subsection{Intent Classification Dataset: CATSLU}
CATSLU dataset is the dataset used in the 1st Chinese Audio-Textual Spoken Language Understanding Challenge. However, there was no previous work on the audio-only portion of the CATSLU dataset, and the competitions are mostly designed for semantic predictions (slot filling). A reference model provided by the CATSLU challenge on the audio-only portion indeed produced relatively poor results (accuracy below 30\%). In addition, the CATSLU dataset is quite imbalanced, as shown in Table \ref{tab:catslu-original}.

 To restructure the dataset into a form suitable for intent classification, we shifted part of the test data into the training set, removed ambiguous utterances, and merged entries from different domains. Each domain in the original data is assigned as an intent. The final distribution by label of the training data is shown in Table \ref{tab:catslu-label}. 

\begin{table}
\centering
\begin{tabular}{ |c|c|c|c| } 
 \hline
 Domain & \multicolumn{3}{|c|}{\# of utterances} \\
 \cline{2-4}
                        & Train & Dev & Test \\
 \hline
  Map       & 5093 & 921 & 1578 \\ 
 \hline
  Music     & 2189 & 381 & 676 \\ 
 \hline
  Weather   &  341 & 378 & 2660 \\ 
 \hline
  Video     &  205 & 195 & 1641  \\ 
 \hline
\end{tabular}
\vspace*{+2mm}
\caption{\label{tab:catslu-original}CATSLU: Original data distribution.}
\end{table}

\begin{table}
\centering
\begin{tabular}{ |c|c|c|c| } 
 \hline
 Label & \multicolumn{3}{|c|}{\# of utterances} \\
 \cline{2-4}
             & Train & Dev & Test \\
 \hline
  Navigation & 2934 & 666 & 1109 \\ 
 \hline
  Music      & 1524 & 251 & 463 \\ 
 \hline
  Weather    & 1463 & 211 & 417 \\ 
 \hline
  Video      & 1004 & 163 & 487  \\ 
 \hline
\end{tabular}
\vspace*{+2mm}
\caption{\label{tab:catslu-label}CATSLU: Distribution of shifted and cleaned data.}
\end{table}

\subsubsection{Frequent Phones}

\begin{figure}
   \includegraphics[width=1.0\linewidth]{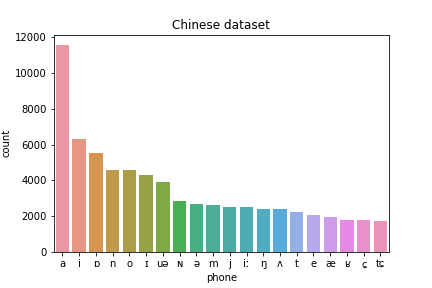}
  \caption{\label{fig:phone} Distribution of phones in the intent classification datasets}
\end{figure}

\begin{figure}
   \includegraphics[width=1.0\linewidth]{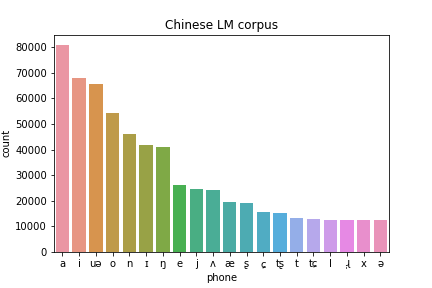}
  \caption{\label{fig:phone_corpus} Distribution of phones in the corpus for pre-training}
\end{figure}

Here we provide distribution of top-20 frequent phones in the pre-training LM corpus and intent classification dataset in Figure \ref{fig:phone} and Figure \ref{fig:phone_corpus} respectively.
We found that they share a large portion of frequent phones. For example, phones like \texttt{a, i, o, n} appear as most frequent phones in both datasets.

\subsubsection{Utterance Lengths}

\begin{figure}[!ht]
\centering
   \includegraphics[width=0.7\linewidth]{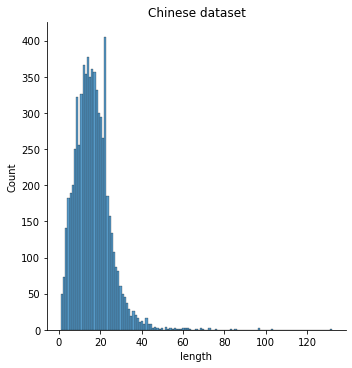}
  \caption{\label{fig:lens} Distribution of utterance lengths in the intent classification datasets}
\end{figure}

\begin{figure}
\centering
   \includegraphics[width=0.7\linewidth]{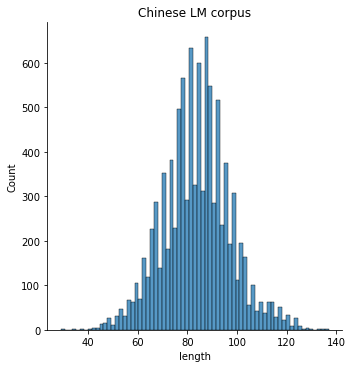}
  \caption{\label{fig:lens_corpus} Distribution of utterance lengths in the corpus for pre-training}
\end{figure}

We also provide the distribution of utterance lengths in Figure \ref{fig:lens} and Figure \ref{fig:lens_corpus}. As shown in the pictures, the LM dataset has better distribution in length with varying lengths in utterances, whereas the intent classification dataset mainly consists of short utterances. The difference in the average length of utterances is illustrated in Table \ref{tab:lens}.

\begin{table}
\centering
\begin{tabular}{ |c|c|c| } 
 \hline
  & Intent Classification  &  LM Corpus  \\ 
 \hline
 Average  & 16.47 & 85.84  \\ 
 \hline
\end{tabular}
\caption{\label{tab:lens} Average utterance lengths for both datasets }
\end{table}

\section{Baseline Results}

\begin{table}[!ht]
\centering
\begin{tabular}{|l|l|l|l|}
\hline
Classes   & Precision & Recall & F1   \\ \hline
0 (Navigation)        & 0.74      & 0.72   & 0.73    \\ \hline
1 (Music)        & 0.47      & 0.56   & 0.51     \\ \hline
2 (Video)        & 0.65      & 0.34   & 0.45      \\ \hline
3 (Weather)        & 0.57      & 0.82   & 0.68      \\ \hline
Macro Avg & 0.61      & 0.61   & 0.59   \\ \hline
Accuracy  & \multicolumn{3}{l|}{0.63}    \\ \hline
\end{tabular}
\caption{\label{tab:base-cn}Baseline (CNN + LSTM) results on the test set.}
\end{table}

The baseline results on the test set are shown in Table \ref{tab:base-cn}.
For the baseline model (CNN + LSTM), we obtain an accuracy 63\% on the intent classification dataset.
In addition, we perform various experiments on the baseline architecture on the development set to see if they will give accuracy improvement. Experiments include increasing the hidden dimension of LSTM (e.g. to 256, 512, 1024), increasing the number of LSTM layers, increasing the embedding dimension, and increasing the number of filters or the filter size for CNN.
However, we found that they do not give much accuracy improvement and will degrade the performance in many cases.

\section{Language Model}

We present a transformer-based language model with the masked language modelling pre-training objective used in RoBERTa. We summarize our results in the following sections.

\subsection{Results without Pre-training}

We first report our results without pre-training in Table \ref{tab:bert-cn}. 
In this setting, we directly train our model for the intent classification task without doing the masked language modelling (MLM) pre-training task.
Our model achieves a F1 score of 53\% and accuracy of 61\% on the intent classification dataset. The performance of our model is comparable to that of the baseline.

The intent classification dataset has both audio data and text data. Besides training phone-based RoBERTa whose input is phone transcript, we also finetune a pretrained text-based BERT model \footnote{https://huggingface.co/bert-base-chinese} whose input is text transcript. The text-based BERT model should provide an upper limit for the phone-based RoBERTa. The performance of the text-based BERT model is summarized in Table \ref{tab:bert-text}.

\begin{table}
\centering
\begin{tabular}{|l|l|l|l|}
\hline
Classes   & Precision & Recall & F1   \\ \hline
0 (Navigation)        & 0.61      & 0.87   & 0.72     \\ \hline
1 (Music)        & 0.53      & 0.24   & 0.33      \\ \hline
2 (Video)        & 0.60      & 0.29   & 0.39     \\ \hline
3 (Weather)        & 0.66      & 0.70   & 0.68     \\ \hline
Macro Avg & 0.60      & 0.52   & 0.53  \\ \hline
Accuracy  & \multicolumn{3}{l|}{0.61}    \\ \hline
\end{tabular}
\caption{\label{tab:bert-cn}RoBERTa results on the test set.}
\end{table}

\begin{table}
\centering
\begin{tabular}{|l|l|l|}
\hline
                & F1 & Accuracy \\ \hline
text-based BERT & 0.91  & 0.93     \\ \hline
\end{tabular}
\caption{\label{tab:bert-text}BERT results on the test set.}
\end{table}

\subsection{Pre-training}

\begin{table}
\centering
\begin{tabular}{|l|l|l|}
\hline
Model & F1 & Accuracy \\ \hline
RoBERTa & 0.53 & 0.61 \\
Pretrained RoBERTa & 0.55 & 0.61 \\ \hline
Baseline & 0.59 & 0.63 \\
Pretrained Baseline & 0.64 & 0.68 \\ \hline
\end{tabular}
\caption{\label{tab:bert-pre}Performance differences on the test set with pre-training.}
\end{table}

In order to improve the performance of our model on the intent classification dataset, we also attempted pre-training RoBERTa. Given that there are around 10GB-12GB audio data in the pre-training corpus, we believe the corpus is sufficiently representative of the language. Our finding, summarized in Table \ref{tab:bert-pre}, shows that pre-training does contribute to the performance on the intent classification task. Both the baseline model and RoBERTa get better performance on the downstream task when they are pre-trained. The performance boost can be as large as 5\% in accuracy and F1-score.

\subsection{Discussion}

Table \ref{tab:bert-pre} compares the performance of RoBERTa against the baseline model. RoBERTa performs slightly worse than the baseline model. We argue that the worse performance of RoBERTa can be explained by the lack of data in the intent classification dataset. RoBERTa has more parameters than the baseline model and should require more training examples to gain a deep understanding of the input phones and the underlying task. For the navigation intent, which has more training examples than other intents as shown in Table \ref{tab:catslu-label}, RoBERTa achieves almost the same performance as the baseline model, as shown in Table \ref{tab:base-cn} and \ref{tab:bert-cn}. 

\section{Conclusion}

In this paper, we propose a novel transformer-based acoustic-only SLU system. Our work is the first attempt to apply BERT-based model on phonetic transcripts. Our work is also the first attempt to use the phonetic units from natural speech for intent classification.
We test both the previous method (CNN + LSTM) and our system on the CATSLU dataset, which contains natural speech instructions in smart light domain and smart assistant domain respectively. We found our system shows reasonable performance, though slightly worse than the SOTA model due to the data-demanding nature of BERT-based models. In addition, we found pre-training improves both the performances of both the previous method and our system. Future work may replicate our general SLU pipeline, and further fine-tune the architecture on larger intent classifications datasets as well as low-resourced language datasets.

\bibliography{anthology,custom}

\begin{thebibliography}{10}
\expandafter\ifx\csname natexlab\endcsname\relax\def\natexlab#1{#1}\fi

\bibitem[{Devlin et~al.(2019)Devlin, Chang, Lee, and
  Toutanova}]{devlin-etal-2019-bert}
Jacob Devlin, Ming-Wei Chang, Kenton Lee, and Kristina Toutanova. 2019.
\newblock \href {https://doi.org/10.18653/v1/N19-1423} {{BERT}: Pre-training of
  deep bidirectional transformers for language understanding}.
\newblock In \emph{Proceedings of the 2019 Conference of the North {A}merican
  Chapter of the Association for Computational Linguistics: Human Language
  Technologies, Volume 1 (Long and Short Papers)}, pages 4171--4186,
  Minneapolis, Minnesota. Association for Computational Linguistics.

\bibitem[{Dong~Wang(2015)}]{THCHS30_2015}
Zhiyong~Zhang Dong~Wang, Xuewei~Zhang. 2015.
\newblock \href {http://arxiv.org/abs/1512.01882} {Thchs-30 : A free chinese
  speech corpus}.

\bibitem[{Gupta et~al.(2021)Gupta, Li, Rallabandi, and Black}]{akshat2021}
Akshat Gupta, Xinjian Li, Sai~Krishna Rallabandi, and Alan~W Black. 2021.
\newblock \href {https://doi.org/10.1109/ICASSP39728.2021.9415112} {Acoustics
  based intent recognition using discovered phonetic units for low resource
  languages}.
\newblock In \emph{ICASSP 2021 - 2021 IEEE International Conference on
  Acoustics, Speech and Signal Processing (ICASSP)}, pages 7453--7457.

\bibitem[{Gupta et~al.(2020)Gupta, Rallabandi, and Black}]{gupta2020mere}
Akshat Gupta, Sai~Krishna Rallabandi, and Alan~W Black. 2020.
\newblock Mere account mein kitna balance hai?--on building voice enabled
  banking services for multilingual communities.
\newblock \emph{arXiv preprint arXiv:2010.16411}.

\bibitem[{Li et~al.(2020)Li, Dalmia, Li, Lee, Littell, Yao, Anastasopoulos,
  Mortensen, Neubig, Black, and Metze}]{allosaurus2020}
Xinjian Li, Siddharth Dalmia, Juncheng Li, Matthew Lee, Patrick Littell, Jiali
  Yao, Antonios Anastasopoulos, David~R. Mortensen, Graham Neubig, Alan~W
  Black, and Florian Metze. 2020.
\newblock \href {https://doi.org/10.1109/ICASSP40776.2020.9054362} {Universal
  phone recognition with a multilingual allophone system}.
\newblock In \emph{ICASSP 2020 - 2020 IEEE International Conference on
  Acoustics, Speech and Signal Processing (ICASSP)}, pages 8249--8253.

\bibitem[{Liu et~al.(2019)Liu, Ott, Goyal, Du, Joshi, Chen, Levy, Lewis,
  Zettlemoyer, and Stoyanov}]{liu2019roberta}
Yinhan Liu, Myle Ott, Naman Goyal, Jingfei Du, Mandar Joshi, Danqi Chen, Omer
  Levy, Mike Lewis, Luke Zettlemoyer, and Veselin Stoyanov. 2019.
\newblock \href {http://arxiv.org/abs/1907.11692} {Roberta: A robustly
  optimized bert pretraining approach}.
\newblock Cite arxiv:1907.11692.

\bibitem[{Lugosch et~al.(2019)Lugosch, Ravanelli, Ignoto, Tomar, and
  Bengio}]{Lugosch2019}
Loren Lugosch, Mirco Ravanelli, Patrick Ignoto, Vikrant~Singh Tomar, and Yoshua
  Bengio. 2019.
\newblock \href {https://doi.org/10.21437/Interspeech.2019-2396} {{Speech Model
  Pre-Training for End-to-End Spoken Language Understanding}}.
\newblock In \emph{Proc. Interspeech 2019}, pages 814--818.

\bibitem[{Radfar et~al.(2020)Radfar, Mouchtaris, and Kunzmann}]{Radfar2020}
Martin Radfar, Athanasios Mouchtaris, and Siegfried Kunzmann. 2020.
\newblock \href {https://doi.org/10.21437/Interspeech.2020-1963} {{End-to-End
  Neural Transformer Based Spoken Language Understanding}}.
\newblock In \emph{Proc. Interspeech 2020}, pages 866--870.

\bibitem[{Vaswani et~al.(2017)Vaswani, Shazeer, Parmar, Uszkoreit, Jones,
  Gomez, Kaiser, and Polosukhin}]{Trans2017}
Ashish Vaswani, Noam Shazeer, Niki Parmar, Jakob Uszkoreit, Llion Jones,
  Aidan~N Gomez, \L~ukasz Kaiser, and Illia Polosukhin. 2017.
\newblock \href
  {https://proceedings.neurips.cc/paper/2017/file/3f5ee243547dee91fbd053c1c4a845aa-Paper.pdf}
  {Attention is all you need}.
\newblock In \emph{Advances in Neural Information Processing Systems},
  volume~30. Curran Associates, Inc.

\bibitem[{Zhu et~al.(2019)Zhu, Zhao, Zhao, Zong, and Yu}]{catslu2019}
Su~Zhu, Zijian Zhao, Tiejun Zhao, Chengqing Zong, and Kai Yu. 2019.
\newblock \href {https://doi.org/10.1145/3340555.3356098} {Catslu: The 1st
  chinese audio-textual spoken language understanding challenge}.
\newblock In \emph{2019 International Conference on Multimodal Interaction},
  ICMI '19, page 521–525, New York, NY, USA. Association for Computing
  Machinery.

\end{thebibliography}
\bibliographystyle{acl_natbib}




\end{document}